\documentclass[11pt,letterpaper]{article}
\usepackage{naaclhlt2016}
\usepackage{times}
\usepackage{latexsym}
\usepackage{graphicx}
\usepackage{amsmath}

\naaclfinalcopy % Uncomment this line for the final submission
\def\naaclpaperid{***} %  Enter the naacl Paper ID here

\title{Comparing Fifty Natural Languages and Twelve Genetic Languages Using Word Embedding Language Divergence (WELD) as a Quantitative Measure of Language Distance}

\author{Ehsaneddin Asgari \and Mohammad R.K. Mofrad \\
Departments of Bioengineering\\
University of California, Berkeley\\
Berkeley, CA 94720, USA\\
\small{\tt asgari@ischool.berkeley.edu, mofrad@berkeley.edu }}

\date{}

\begin{document}

\maketitle

\begin{abstract}
We introduce a new measure of distance between languages based on word embedding, called word embedding language divergence (WELD). WELD is defined as divergence between unified similarity distribution of words between languages. Using such a measure, we perform language comparison for fifty natural languages and twelve genetic languages. Our natural language dataset is a collection of sentence-aligned parallel corpora from bible translations for fifty languages spanning a variety of language families. Although we use parallel corpora, which guarantees having the same content in all languages, interestingly in many cases languages within the same family cluster together. In addition to natural languages, we perform language comparison for the coding regions in the genomes of 12 different organisms (4 plants, 6 animals, and two human subjects). Our result confirms a significant high-level difference in the genetic language model of humans/animals versus plants. The proposed method is a step toward defining a quantitative measure of similarity between languages, with applications in languages classification, genre identification, dialect identification, and evaluation of translations. 
\end{abstract}
\vspace{-0.25cm}
\section{Introduction}
Classification of language varieties is one of the prominent problems in linguistics~\cite{smith2016dynamic}. The term language variety can refer to different styles, dialects, or even a distinct language~\cite{marjorie2001language}. It has been a longstanding argument that strictly quantitative methods can be applied to determine the degree of similarity or dissimilarity between languages~\cite{kroeber1937quantitative,sankaran1950quantitative,kramsky1959quantitative,mcmahon2003finding}. The methods proposed in the 1990's and early 2000' mostly relied on utilization of intensive linguistic resources. For instance, similarity between two languages was defined based on the number of common cognates or phonological patterns according to a manually extracted list~\cite{kroeber1937quantitative,mcmahon2003finding}. Such an approach, of course, is not easily extensible to problems involving new languages. Recently, statistical methods have been proposed to automatically detect cognates ~\cite{berg2010phylogenetic,hall2010finding,bouchard2013automated,CiobanuDinu2014} and subsequently compare  languages based on the number of common cognates~\cite{CiobanuDinu2014}.

In this paper our aim is to define a quantitative measure of distance between languages. Such a metric should reasonably take both syntactic and semantic variability of languages into account. A measure of distance between languages can have various applications including quantitative genetic/typological language classification, styles and genres identification, and translation evaluation. In addition, comparing the biological languages generating the genome in different organisms can potentially shed light on important biological facts.

%In this framework, we consider each language as a graph of words, where words are sharing various degrees of syntactical and semantic similarities with each-other.
\vspace{-0.25cm}
\subsection{Problem Definition}
\label{sect:intro}
Our goal is to be able to provide a quantitative estimate of distance for any two given languages. In our framework, we define a language as a weighted graph $\Omega_L(V,e)$, where $V$ is a set of vertices (words), and $e: (V\times V) \rightarrow \Re$ is a weight function mapping a pair of words to their similarity value. Then our goal of approximating the distance between the two languages $L$ and $L'$ can be transferred to the approximation of the distance between $\Omega_L(V,e)$ and $\Omega_{L'}(V',e')$. In order to approach such a problem firstly we need to address the following questions:
\vspace{-0.25cm}
\begin{itemize}
    \item What is a proper weight function $e$ estimating a similarity measure between words $w_i,w_j \in V$ in a language $L$?
    \vspace{-0.25cm}
    \item How can we relate words in $V$ to words in $V'$?
    \vspace{-0.75cm}
    \item And finally, how can we measure a distance between languages $\Omega_{L}$ and $\Omega_{L'}$, which means $D(\Omega_{L},\Omega_{L'})$?
    \vspace{-0.25cm}
\end{itemize}

In the following section we explain how researchers have addressed the above mentioned questions until now. 

\subsubsection{Word similarity within a language}
\label{sect:wordsim}
The main aim of word similarity methods is to measure how similar pairs of words are to each-other, semantically and syntactically~\cite{han2013improving}. Such a problem has a wide range of applications in information retrieval, automatic speech recognition, word sense disambiguation, and machine translation ~\cite{collobert2008unified,glorot2011domain,mikolov2013linguistic,turney2010frequency,resnik1999semantic,schwenk2007continuous}.

Various methods have been proposed to measure word similarity, including thesaurus and taxonomy-based approaches, data-driven methods, and hybrid techniques~\cite{miller1995wordnet,mohammad2006distributional,mikolov2013efficient,han2013improving}. Taxonomy-based methods are not easily extensible as they usually require extensive human intervention for creation and maintenance~\cite{han2013improving}. One of the main advantages of data-driven methods is that they can be employed even for domains with shortage of manually annotated data. 

Almost all of the data-driven methods such as matrix factorization~\cite{xu2003document}, word embedding~\cite{mikolov2013efficient}, topic models~\cite{blei2012probabilistic}, and mutual information~\cite{han2013improving} are based on co-occurrences of words within defined units of text data. Each method has its own convention for unit of text, which can be a sentence, paragraph or a sliding window around a word. Using distributed representations have been one of the most successful approaches for computing word similarity in natural language processing~\cite{collobert2011natural}. The main idea in distributed representation is characterizing words by the company they keep~\cite{hinton1984distributed,firth1975modes,collobert2011natural}. 

Recently, continuous vector representations known as word vectors have become popular in natural language processing (NLP) as an efficient approach to represent semantic/syntactic units~\cite{mikolov2013efficient,collobert2011natural}. Word vectors are trained in the course of training a language model neural network from large amounts of textual data (words and their contexts)~\cite{mikolov2013efficient}. More precisely, word representations are the outputs of the last hidden layer in a trained neural network for language modeling. Thus, word vectors are supposed to encode the most relevant features to language modeling by observing various samples. In this representation similar words have closer vectors, where similarity is defined in terms of both syntax and semantics. By training word vectors over large corpora of natural languages, interesting patterns have been observed. Words with similar vector representations display multiple types of similarity. For instance, $\overrightarrow{King}-\overrightarrow{Man}+\overrightarrow{Woman}$ is the closest vector to that of the word $\overrightarrow{Queen}$ (an instance of semantic regularities) and $\overrightarrow{quick}- \overrightarrow{quickly} \approx \overrightarrow{slow}- \overrightarrow{slowly}$ (an instance of syntactic regularities). A recent work has proposed the use of word vectors to detect linguistic changes within the same language over time~\cite{kulkarni2015statistically}. The fact that various degrees of similarity were captured by such a representation convinced us to use it as a notion of proximity for words.

\subsubsection{Word alignment}
\label{sect:wordalg}
As we discussed in section~\ref{sect:intro}, in order to compare graphs $\Omega_L$ and $\Omega_L'$, we need to have a unified definition of words (vertices). Thus, we need to find a mapping function from the words in $V$ to the words in $V'$. Obviously when two languages have the same vocabulary set this step can be skipped, which is the case when we perform within-language genres analysis or linguistic drifts study ~\cite{stamatatos2000text,kulkarni2015statistically}, or even when we compare biological languages (DNA or protein languages) for different species~\cite{asgari2015continuous}. However, when our goal is to compare distributional similarity of words for two different languages, such as French and German, we need to find a mapping from words in French to German words. 

Finding a word mapping function between two languages can be achieved using a dictionary or using statistical word alignment in parallel corpora~\cite{och2003systematic,lardilleux2009sampling}. Statistical word alignment is a vital component in any statistical machine translation pipeline~\cite{fraser2007measuring}. Various methods/tools has been proposed for word alignment, such as GIZA++~\cite{och2003giza} and Anymalign~\cite{lardilleux2009sampling}, which are able to extract high quality word alignments from sentence-aligned multilingual parallel corpora.

One of the data resources we use in this project is a large collection of sentence-aligned parallel corpora we extract from bible translations in fifty languages. Thus, in order to find a word mapping function among all these languages we used statistical word alignment techniques and in particular Anymalign~\cite{lardilleux2009sampling}, which can process any number of languages at once.

\subsubsection{Network Analysis of Languages}
\label{sect:langnet}
The rather intuitive approach of treating languages as networks of words has been proposed and explored in the last decade by a number of researchers~\cite{i2001small,liu2013language,cong2014approaching,gao2014comparison}. In these works, human languages, like many other aspects of human behavior, are modeled as complex networks~\cite{costa2011analyzing}, where the nodes are essentially the words of the language and the weights on the edges are calculated based on the co-occurrences of the words~\cite{liu2013language,i2001small,gao2014comparison}. Clustering of 14 languages based on various parameters of a complex network such as average degree, average path length, clustering coefficient, network centralization, diameter, and network heterogeneity has been done by~\cite{liu2013language}. A similar approach is suggested by~\cite{gao2014comparison} for analysis of the complexity of six languages. Although, all of the above mentioned methods have presented promising results about similarity and regularity of languages, to our understanding they need the following improvements:

\textbf{Measure of word similarity:} Considering co-occurrences as a measure of similarity between nodes, which is the basis of the above mentioned complex network methods, is a naive estimate of similarity,~\cite{liu2013language,i2001small,gao2014comparison}. The most trivial cases are synonyms, which we expect to be marked as the most similar words to each other. However, since they can only be used interchangeably with each other in the same sentences, their co-occurrences rate is very low. Thus, raw co-occurrence is not necessarily a good indicator of similarity.

\textbf{Independent vs. joint analysis:} Previous methods have compared the parameters of language graphs independently, except for some relatively small networks of words for illustration~\cite{liu2013language,i2001small,gao2014comparison}. However, two languages may have similar settings of the edges but for completely different concepts. Thus, a systematic way for joint comparison of these networks is essential.

\textbf{Language collection:} The previous analysis was performed on a relatively small number of languages. For instance in~\cite{liu2013language}, fourteen languages were studied where twelve of them were from the Slavic family of languages, and ~\cite{gao2014comparison} studied six languages. Clearly, studying more languages from a broader set of language families would be more indicative.
\vspace{-0.25cm}
\subsection{Our Contributions}
In this paper, we suggest a heuristic method toward a quantitative measure of distance between languages. We propose divergence between unified similarity distribution of words as a quantitative measure of distance between languages.

\textbf{Measure of word similarity:} We use cosine similarity between word vectors as the metric of word similarities, which has been shown to take into account both syntactic and semantic similarities~\cite{mikolov2013efficient}. Thus, in the weighted language graph $\Omega_L(V,e)$, the weight function $e: (V\times V) \rightarrow \Re$ is defined by word-vector cosine similarities between pairs of words. Although word vectors are calculated based on co-occurrences of words within sliding windows, they are capable of attributing a reasonable degree of similarity to close words that do not co-occur.

\textbf{Joint analysis of language graphs:} By having word vector proximity as a measure of word similarity, we can represent each language as a joint similarity distribution of its words. Unlike the methods mentioned in section~\ref{sect:langnet} which focused on network properties and did not consider a mapping function between nodes across various languages, we propose performing node alignment between different languages~\cite{lardilleux2009sampling}. Consequently, calculation of Jensen-Shannon divergence between unified similarity distributions of the languages can provide us with a measure of distance between languages.

\textbf{Language collection:} In this study we perform language comparison for fifty natural languages and twelve genetic language. 

\textit{Natural languages:} We extracted a collection of sentence-aligned parallel corpora from bible translations for fifty languages spanning a variety of language families including Indo-European (Germanic, Italic, Slavic, Indo-Iranian), Austronesian, Sino-Tibetan, Altaic, Uralic, Afro-Asiatic, etc. This set of languages is relatively large and diverse in comparison with the corpora that have been used in previous studies~\cite{liu2013language,gao2014comparison}. We calculated the Jensen-Shannon divergence  between joint similarity distributions for fifty language graphs consisting of 4,097 sets of aligned words in all these fifty languages. Using the mentioned divergence we performed cluster analysis of languages. Interestingly in many cases languages within the same family clustered together. In some cases, a lower degree of divergence from the source language despite belonging to different language families was indicative of a consistent translation. 

\textit{Genetic languages:}
Nature uses certain languages to generate biological sequences such as DNA, RNA, and proteins. Biological organisms use sophisticated languages to convey information within and between cells, much like humans adopt languages to communicate~\cite{yandell2002genomics,searls2002language}. Inspired by this conceptual analogy, we use our languages comparison method for comparison of genetic languages in different organisms. Genome refers to a sequence of nucleotides containing our genetic information. Some parts of our genome are coded in a way that can be translated to proteins (exonic regions), while some regions cannot be translated into proteins (introns)~\cite{saxonov2000eid}. In this study, we perform language comparison of coding regions in 12 different species (4 plants, 6 animals, and two human subjects). Our language comparison method is able to assign a reasonable relative distance between species.
\vspace{-0.25cm}
\section{Methods}
\label{sect:method}
\vspace{-0.25cm}
As we discussed in~\ref{sect:intro}, we transfer the problem of finding a measure of distance between languages $L$ and $L'$ to finding the distance between their language graphs $\Omega_L(V,e)$ and $\Omega_{L'}(V',e')$.

\textbf{Word Embedding:} We define the edge weight function $e: (V\times V) \rightarrow \Re$ to be the cosine similarity between word vectors.

\textbf{Alignment:} When two languages have different words, in order to find a mapping between the words in $V$ and $V'$ we can perform statistical word alignment on parallel corpora.

\textbf{Divergence Calculation:} Calculating Jensen-Shannon divergence between joint similarity distributions of the languages can provide us with a notion of distance between languages.

Our language comparison method has three components. Firstly, we need to learn word vectors from large amounts of data in an unsupervised manner for both of the languages we are going to compare. Secondly, we need to find a mapping function for the words and finally we need to calculate the divergence between languages. In the following section we explain each step aligned with the experiment we perform on both natural languages and genetic languages.
\vspace{-0.25cm}
\subsection{Learning Word Embedding}
\label{sect:embed}
Word embedding can be trained in various frameworks (e.g. non-negative matrix factorization and neural network methods~\cite{mikolov2013linguistic,levy2014neural}). Neural network word embedding trained in the course of language modeling is shown to capture interesting syntactic and semantic regularities in the data~\cite{mikolov2013linguistic,mikolov2013efficient}. Such word embedding known as word vectors need to be trained from a large number of training examples, which are basically words and their corresponding contexts. In this project, in particular we use an implementation of the skip-gram neural network~\cite{mikolov2013distributed}.

In training word vector representations, the skip-gram neural network attempts to maximize the average probability of contexts for given words in the training data: 

\begin{equation}
\label{eq:skipgram}
\begin{split}
&\underset{v,v'}{\mathrm{argmax}} \frac{1}{N}\sum_{i=1}^{N}\sum_{-c \leq j \leq c, j\neq 0 }\log{p(w_{i+j}|w_i)}\\
&p(w_{i+j}|w_i)=\frac{\exp{({{{v'}^T_{w_{i+j}}}} v_{w_{i}})}}{\sum_{k=1}^{W}\exp{({{{v'}^T_{w_{k}}}} v_{w_{i}})}},
\end{split}
\end{equation}

where $N$ is the length of the training, $2c$ is the window size we consider as the context, $w_i$  is the center of the window, $W$ is the number of words in the dictionary and $v_w$ and $v'_w$ are the n-dimensional word representation and context representation of word $w$, respectively. At the end of the training the average of $v_w$ and $v'_w$ will be considered as the word vector for $w$. The probability $p(w_{i+j}|w_i)$ is defined using a softmax function. In the implementation we use (Word2Vec)~\cite{mikolov2013distributed} negative sampling has been utilized, which is considered as the state-of-the-art for training word vector representation. 

\subsubsection{Natural Languages Data}
\label{sect:data}
For the purpose of language classification we need parallel corpora that are translated into a large number of languages, so that we can find the alignments using statistical methods. Recently, a massive parallel corpus based on 100 translations of the Bible has been created in XML format~\cite{christodouloupoulos2015massively}, which we choose as the database for this project. In order to make sure that we have a large enough corpus for learning word vectors, we pick the languages for which translations of both the Old Testament and the New Testament are available. From among those languages we pick the ones containing all the verses in the Hebrew version (which is the source language for most of the data) and finally we end up with almost 50 languages, containing 24,785 aligned verses. For Thai, Japanese, and Chinese we use the tokenized versions in the database~\cite{christodouloupoulos2015massively}. In addition, before feeding the skip-gram neural network we remove all punctuation.

In our experiment, we use the word2vec implementation of skip-gram~\cite{mikolov2013distributed}. We set the dimension of word vectors $d$ to 100, and the window size $c$ to 10 and we sub-sample the frequent words by the ratio $\frac{1}{10^3}$.

\subsubsection{Genetic Languages Data}
In order to compare the various genetic languages we use the IntronExon database that contains coding and non-coding regions of genomes for a number of organisms~\cite{shepelev2006advances}. From this database we extract a data-set of coding regions (CR) from 12 organisms consisting of 4 plants (arabidopsis, populus, moss, and rice), 6 animals (sea-urchin, chicken, cow, dog, mouse, and rat), and two human subjects. The number of coding regions we have in the training data for each organism is summarized in Table~\ref{genomelang-table}. The next step is splitting each  sequence to a number of words. Since the genome is composed of the four DNA nucleotides A,T,G and C, if we split the sequences in the character level the language network would be very small. We thus split each sequence into n-grams ($n=3,4,5,6$), which is a common range of n-grams in bioinformatics\cite{ganapathiraju2002comparative,mantegna1995systematic}. As suggested by\cite{asgari2015continuous} we split the sequence into non-overlapping n-grams, but we consider all possible ways of splitting for each sequence.

\begin{table}[ht!]
\small
\centering
\scalebox{0.8}{\begin{tabular}{|l|rl|}
\hline \bf Organisms & \bf \# of CR & \bf  \# of 3-grams \\ \hline
Arabidopsis &  179824 & 42,618,288  \\
Populus &  131844 & 28,478,304  \\
Moss & 167999 & 38,471,771 \\
Rice & 129726 & 34,507,116 \\
Sea-urchin & 143457 & 27,974,115  \\
Chicken &   187761 & 34,735,785\\
Cow & 196466 & 43,222,520 \\
Dog & 381147 & 70,512,195  \\
Mouse & 215274 & 34,874,388 \\
Rat &  190989 & 41,635,602 \\
Human 1 & 319391 & 86,874,352 \\
Human 2 & 303872 & 77,791,232 \\
\hline
\end{tabular}}
\caption{\label{genomelang-table} The genome data-set for learning word vectors in different organisms. The number of coding regions and the total occurrences of 3-grams are presented. Clearly, the total number of all n-grams (n=3,4,5,6) is almost the same.\vspace{-0.5cm}}
\end{table}

We train the word vectors for each setting of n-grams and organisms separately, again using skip-gram neural network implementation~\cite{mikolov2013distributed}. We set the dimension of word vectors $d$ to 100, and window size of $c$ to 40. In addition, we sub-sample the frequent words by the ratio $10^-3$.

\subsection{Word Alignment}
\label{sect:align}
The next step is to find a mapping between the nodes in $\Omega_L(V,e)$ and $\Omega_{L'}(V',e')$. Obviously in case of quantitative comparison of styles within the same language we do not need to find an alignment between the nodes in $V$ and $V'$. However, when we are comparing two distinct languages we need to find a mapping from the words in language $L$ to the words in language $L'$. 
\vspace{-0.25cm}
\subsubsection{Word Alignment for Natural Languages}
\label{sect:nlalign}
As we mentioned in section~\ref{sect:data}, our parallel corpora contain texts in fifty languages from a variety of language families. We decided to use statistical word alignments because we already have parallel corpora for these languages and therefore performing statistical alignment is straightforward. In addition, using statistical alignment we hope to see evidences of consistent/inconsistent translations. 

We use an implementation of Anymalign~\cite{lardilleux2009sampling}, which is designed to extract high quality word alignments from sentence-aligned multilingual parallel corpora. Although Anymalign is capable of performing alignments in several languages at the same time, our empirical observation was that performing alignments for all languages against a single language and then finding the global alignment through that alignment is faster and results in better alignments. We thus align all translations with the Hebrew version. To ensure the quality of alignments we apply a high threshold on the score of alignments. In a final step, we combine the results and end up with a set of 4,097 multilingual alignments. Hence we have a mapping from any of the 4,097 words in one language to one in any other given language, where the Hebrew words are unique, but not necessarily the others.

\subsubsection{Genetic Languages Alignment}
In genetic language comparison, since the n-grams are generated from the same nucleotides (A,T,C,G), no alignment is needed and $V$ would be the same as $V'$.

\subsection{Calculation of Language Divergence}
In section~\ref{sect:embed} we explained how to make language graphs $\Omega_L(V,e)$ and $\Omega_{L'}(V',e')$. Then in section~\ref{sect:align} we proposed a statistical alignment method to find the mapping function between the nodes in $V$ and $V'$. Having achieved the mapping between the words in $V$ and the words in $V'$, the next step is comparison of $e$ and $e'$.

In comparing language graphs what is more crucial is the \textit{relative} similarities of words. Intuitively we know that the relative similarities of words vary in different languages due to syntactic and semantic differences. Hence, we decided to use the divergence between relative similarities of words as a heuristic measure of the distance between two languages. To do so, firstly we normalize the relative word vector similarities within each language. Then, knowing the mapping between words in $V$ and $V'$ we unify the coordinates of the normalized similarity distributions. Finally, we calculate the Jensen-Shannon divergence between the normalized and unified similarity distributions of two languages:
\vspace{-0.25cm}
\[
D_{L,L'}=JSD(\hat{e},\hat{e'}),\vspace{-0.25cm}
\]
where $\hat{e}$ and $\hat{e'}$ are normalized and unified similarity distributions of word pairs in $\Omega_L(V,e)$ and $\Omega_{L'}(V',e')$ respectively.

\subsubsection{Natural Languages Graphs}
For the purpose of language classification we need to find pairwise distances between all of the fifty languages we have in our corpora. Using the mapping function obtained from statistical alignments of Bible translations, we produce the normalized and unified similarity distributions of word pairs $\hat{e^{(k)}}$ for language $L^{(k)}$. Therefore to compute the quantitative distance between two languages $L^{(i)}$ and $L^{(j)}$ we calculate $D_{L_i,L_j}=JSD(\hat{e^{(i)}},\hat{e^{(j)}})$.

Consequently, we calculate a quantitative distance between each pair of languages. In a final step, for visualization purposes, we perform Unweighted Pair Group Method with Arithmetic Mean (UPGMA) hierarchical clustering on the pairwise distance matrix of languages~\cite{johnson1967hierarchical}. 

\subsubsection{Genetic Languages Graphs}
The same approach as carried out for natural languages is applied to genetic languages corpora. Pairwise distances of genetic languages were calculated using Jensen-Shannon divergence between normalized and unified similarity distributions of word pairs for each pair of languages. 

We calculate the pairwise distance matrix of languages for each n-gram separately to verify which length of DNA segment is more discriminative between different species. 
\vspace{-0.25cm}
\section{Results}
\label{sect:res}
\vspace{-0.15cm}
\subsection{Classification of Natural Languages}
\begin{figure*}[pt!]
\center{
  \includegraphics[trim={0cm 0cm 0cm 3cm},width=1.0\textwidth]{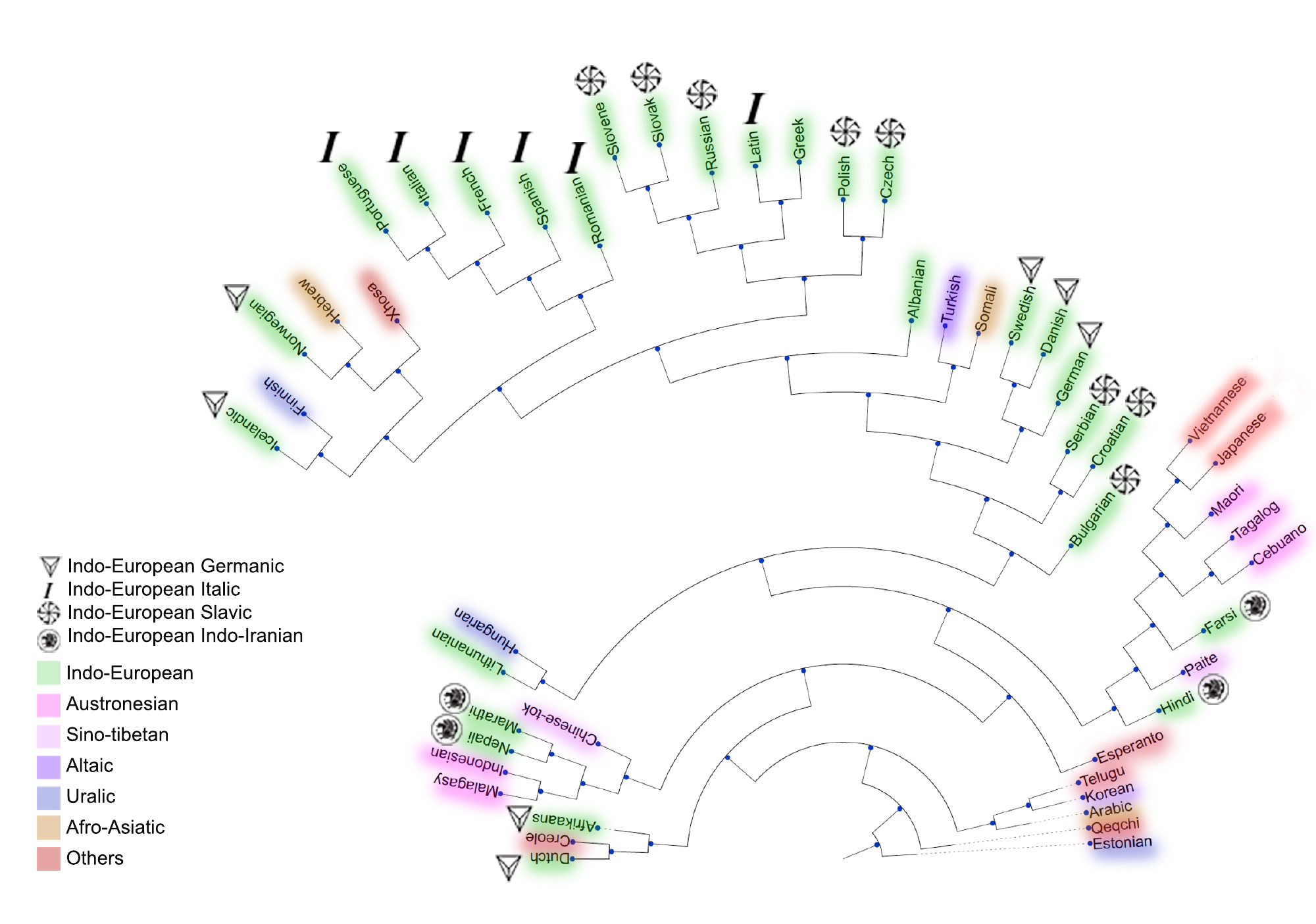}
  \caption{\label{bible-figure} Hierarchical clustering of fifty natural languages according to divergence of joint distance distribution of 4097 aligned words in bible parallel corpora. Subsequently we use colors to show the ground-truth about family of languages. For Indo-European languages we use different symbols to distinguish various sub-families of Indo-European languages. We observe that the obtained clustering reasonably discriminates between various families and subfamilies.}}
\end{figure*}

The result of the UPGMA hierarchical clustering of languages is shown in Figure~\ref{bible-figure}. As shown in this figure, many languages are clustered together according to their family and sub-family. Many Indo-European languages (shown in green) and Austronesian languages (shown in pink) are within a close proximity. Even the proximity between languages within a sub-family are preserved with our measure of language distance. For instance, Romanian, Spanish, French, Italian, and Portuguese, all of which belong to the Italic sub-family of Indo-European languages, are in the same cluster. Similarly, the Austronesian langauges Cebuano, Tagalog, and Maori as well as Malagasy and Indonesian are grouped together.

Although the clustering based on word embedding language divergence matches the genetic/typological classification of languages in many cases, for some pairs of languages their distance in the clustering does not make any genetic or topological sense. For instance, we expected Arabic and Somali as Afro-Asiatic languages to be within a close proximity with Hebrew. However, Hebrew is matched with Norwegian, a Germanic Indo-European language. After further investigations and comparing word neighbors for several cases in these languages, it turns out that the Norwegian bible translation highly matches Hebrew because of being a consistent and high-quality translation. In this translation, synonym were not used interchangeably and language usage stays more faithful to the structure of the Hebrew text.
\vspace{-0.25cm}
\subsubsection{Divergence between Genetic Languages}
\begin{figure*}[pt!]
\center{%trim={20cm 0cm 20cm 0cm},
  \includegraphics[trim={10cm 2cm 10cm 5cm},width=0.65\textwidth]{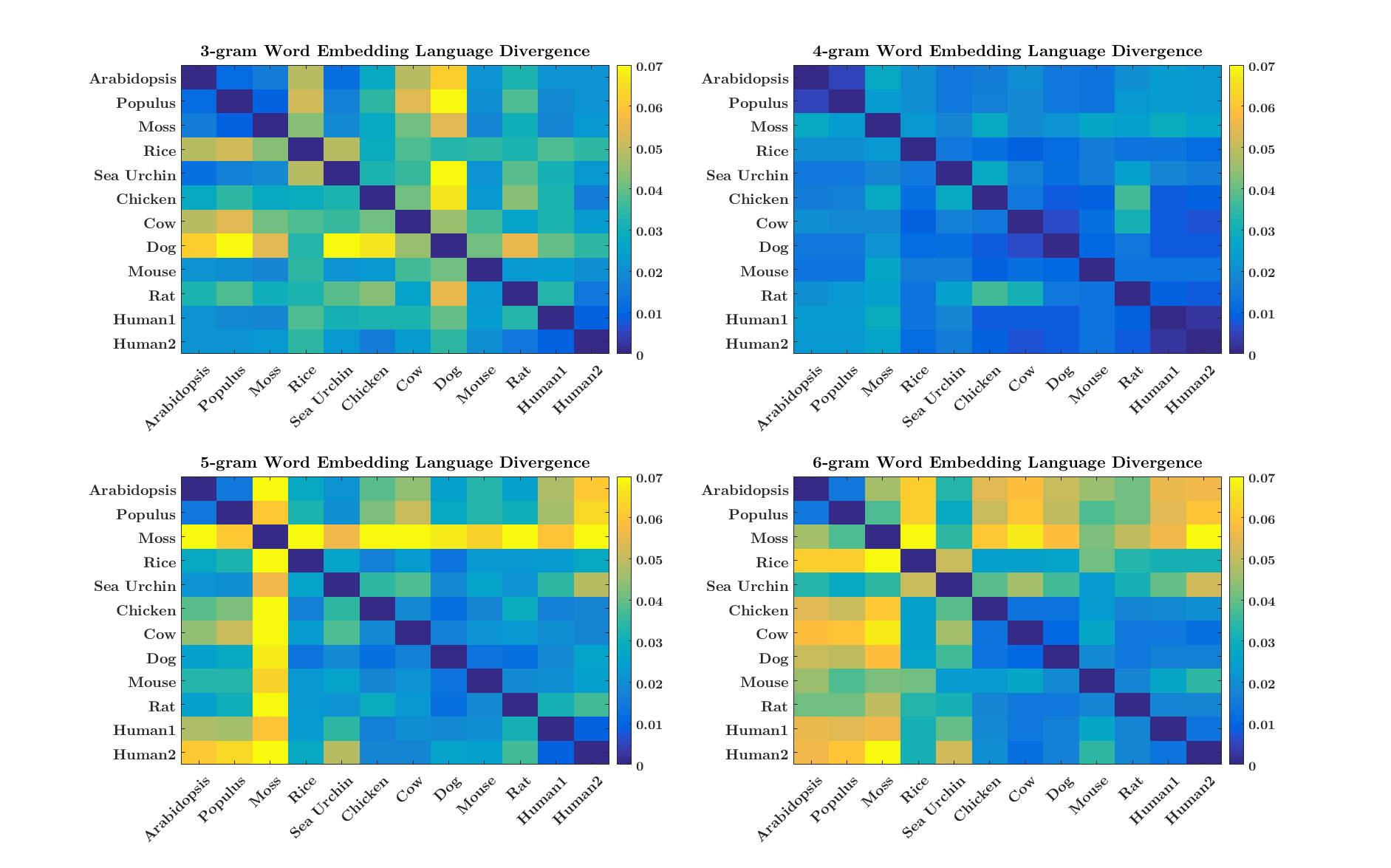}
  \caption{\label{fig:dnalanguage} Visualization of word embedding language divergence in twelve different genomes belonging to 12 organisms for various n-gram segments. Our results indicate that evolutionarily closer species have higher proximity in the syntax and semantics of their genomes.}}
\end{figure*}

The pairwise distance matrix of the twelve genetic languages for  n-grams ($n=3,4,5,6$) is shown in Figure~\ref{fig:dnalanguage}. Our results confirm that evolutionarily closer species have a reasonably higher level of proximity in their language models. We can observe in Figure~\ref{fig:dnalanguage}, that as we increase the number of n-grams the distinction between animal/human genome and plant genome increases.
\vspace{-0.15cm}
\section{Conclusion}
\vspace{-0.25cm}
In this paper, we proposed Word Embedding Language Divergence (WELD) as a new heuristic measure of distance between languages. Consequently we performed language comparison for fifty natural languages and twelve genetic languages. Our natural language dataset was a collection of sentence-aligned parallel corpora from bible translations for fifty languages spanning a variety of language families. We calculated our word embedding language divergence for 4,097 sets of aligned words in all these fifty languages. Using the mentioned divergence we performed cluster analysis of languages. 

The corpora for all of the languages but one consisted of translated text instead of original text in those languages. This means many of the potential relations between words such as collocations and culturally influenced semantic connotations did not have the full chance to contribute to the measured language distances. This can potentially make it harder for the algorithm to detect related languages. In spite of this, however in many cases languages within the same family/sub-family clustered together. In some cases, a lower degree of divergence from the source language despite belonging to different language families was indicative of a consistent translation. This suggests that this method can be a step toward defining a quantitative measure of similarity between languages, with applications in languages classification, genres identification, dialect identification, and evaluation of translations. 

In addition to the natural language data-set, we performed language comparison of n-grams in coding regions of the genome in 12 different species (4 plants, 6 animals, and two human subjects). Our language comparison method confirmed that evolutionarily closer species are closer in terms of genetic language models. Interestingly, as we increase the number of n-grams the distinction between genetic language in animals/human versus plants increases. This can be regarded as indicative of a high-level diversity between the genetic languages in plants versus animals.

\textbf{Acknowledgments}\\
Fruitful discussions with David Bamman, Meshkat Ahmadi, and Mohsen Mahdavi are gratefully acknowledged.

\bibliography{naaclhlt2016}
\bibliographystyle{naaclhlt2016}
\end{document}